\documentclass[11pt]{article}
\usepackage{coling2020}
\usepackage{times}
\usepackage{url}
\usepackage{latexsym}
\usepackage[hidelinks]{hyperref}
\usepackage{makecell,multirow}
\usepackage{cleveref}
\crefformat{section}{\S#2#1#3} 
\crefformat{subsection}{\S#2#1#3}
\crefformat{subsubsection}{\S#2#1#3}

\colingfinalcopy 

\title{Joint Persian Word Segmentation Correction\\and Zero-Width Non-Joiner Recognition Using BERT}

\author{Ehsan Doostmohammadi$^\star$, Minoo Nassajian$^\star$, Adel Rahimi$^\dagger$ \\
  $^\star$Sharif University of Technology, Tehran, Iran \\
  $^\dagger$Dathena Science Pte. Ltd., Singapore \\
  {\tt \{e.doostm72,m.nassajian2016\}@student.sharif.edu} \\
  \tt adel.rahimi@dathena.io\\}  
 
\date{}

\begin{document}
\maketitle

\begin{abstract}
    Words are properly segmented in the Persian writing system; in practice, however, these writing rules are often neglected, resulting in single words being written disjointedly and multiple words written without any white spaces between them.
    This paper addresses the problems of word segmentation and zero-width non-joiner (ZWNJ) recognition in Persian, which we approach jointly as a sequence labeling problem.
    We achieved a macro-averaged F\textsubscript{1}-score of 92.40\% on a carefully collected corpus of 500 sentences with a high level of difficulty.
\end{abstract}

\section{Introduction}
\label{intro}

People who have worked with real-world data in Persian natural language processing (NLP) are familiar with the Persian writing system's problematic properties, the most important of which are word segmentation and ZWNJ, both consequences of a phenomenon called ``orthographic ligature".
In Persian, white space character (\texttt{U+0020}) is used to segment words in a sentence, and ZWNJ (\texttt{U+200C}) is used between free morphemes in compound words (or combining forms), and also between one or several free morphemes and one or several bound morphemes to stop joining letters from connecting into a ligature \cite{saffari86}.
Although there is a set of rules regarding the use of white space and ZWNJ set by the regulatory body of the Persian language \cite{persianort}, only a few people follow them in writing formal Persian, let alone the informal language.

Some common errors regarding the use of white space and ZWNJ are as follows:
(1) not segmenting words where the last letter of the first word is a non-joiner character (e.g., \emph{va␣\={a}m\={a}r[-e]␣marg$|$omir␣behtar␣az␣xeyli$|$h\={a}st}\footnote{In this example from twitter there was no white space or ZWNJ. We added ␣ where white space and $|$ where ZWNJ should have been.} ``and the mortality rate is better than many");
(2) using ZWNJ instead of white space which could be categorized as (a) excessive use of ZWNJ (e.g., for separating first and last names) and (b) ZWNJ key hit mistakenly as in many cell phone keyboards the space key and the ZWNJ key are adjacent to each other;
(3) using space instead of ZWNJ (e.g. \emph{mi konam} ``I do");
(4) using neither of them (e.g. \emph{mikonam} ``I do");
and many other spontaneous errors.

In comparison to European languages, such as English and German, word segmentation in Asian languages like Japanese, Chinese \cite{li2019word}, and Thai \cite{aroonmanakun2002collocation,tesprasit2003learning}, is more complicated, because space is not specifically used as an orthographic word boundary delimiter \cite{khan2018supervised}.
For instance, in Vietnamese, space is not only used to separate words, but it is also applied to separate syllables (that can be considered as a meaningful word or as a part of multi-syllable words) that make up words \cite{huyen2008hybrid}. 
The problem in Persian is not similar to that of the abovementioned languages, but quite identical to Urdu \cite{durrani-hussain-2010-urdu,lehal-2010-word,6473706,bin-zia-etal-2018-urdu,khan2018urdu}, as they both are Indo-Iranian languages and their writing systems are derived from the Arabic script.  In Persian, words are properly segmented in theory, but these rules are not always followed in practice.
These kinds of writing system characteristics result in difficulties in text processing, e.g., sequence labeling.
According to the previous research, word segmentation can improve the results of other NLP tasks like information retrieval \cite{foo2004chinese}, machine translation \cite{xu2004we}, information extraction \cite{peng2016improving}, dependency parsing \cite{nguyen2018neural}, etc.

In this paper, we address the problems of word segmentation and ZWNJ correction in Persian using sequence labeling models and achieve results that are quite promising and could pave the way for an effective solution for real-world situations. To be sure of this, we gathered a corpus of 500 sentences with a high degree of difficulty regarding the correctness of word segmentation and ZWNJs, and evaluated the models' performance on it. After reviewing the related work, we discuss the data used for training, validating, and testing the models in \cref{sec:data}, the methodology and the experimental settings in \cref{sec:meth}, and the results in \cref{sec:res}. We then conclude the paper and suggest some ideas for future work.

\section{Related Work}
\label{sec:rel}

Segmentation techniques could be classified into different categories, such as rule-based methods \cite{riaz2010rule,khan2018urdu}, hybrid methods (\newcite{nakagawa2007hybrid} and \newcite{pham2009hybrid} use a hidden Markov model with some hand-crafted rules), and machine learning and deep learning methods (\newcite{sassano2002empirical} and \newcite{nguyen2006vietnamese} use conditional random fields and support vector machines, and \newcite{ma2018state} use a bidirectional recurrent neural network model). In addition to supervised learning (like \newcite{zhang2016transition},  \newcite{lyu2016joint}, etc.), there are studies on semi-supervised \cite{fujii2017nonparametric,wang2019unsupervised,yang2014semi} and unsupervised word segmentation \cite{mochihashi2009bayesian,seeha2020thailmcut,zhang2018improving,bui2019hmms}. As for the latest studies, we can mention \newcite{ke2020unified}, \newcite{nguyen2018neural}, and \newcite{yan2020graph} that use BERT and multitask learning methods to segment words in Chinese and Vietnamese languages.

In Persian, there are few studies conducted on preprocessing toolkits, some of which also include modules for space and ZWNJ correction. As an early work, we can mention \newcite{shamsfard2009step} that designed a Persian text preprocessing tool called STeP-1 to tokenize texts, check spellings, and analyze words morphologically. To design the tokenizer, they use dictionary-based and rule-based methods to recognize the correct places of space and ZWNJ, and report a performance of 86.6\%.
As another work, \newcite{sarabi2013parsipardaz} design a toolkit for Persian text processing in four lexical, morphological, syntactic, and semantic levels. Space and ZWNJ positions are checked in the first level using dictionary-based and rule-based methods, achieving a precision of 95\%.
Additionally, Hazm\footnote{\href{https://github.com/sobhe/hazm}{https://github.com/sobhe/hazm}}, an open-source preprocessing library, performs ZWNJ correction using regular expressions and a list of valid stems. Parsivar \cite{mohtaj2018parsivar}, the most powerful Persian preprocessing tool, applies rule-based and statistical methods to determine the correct places of ZWNJ and space, respectively. They trained a naive Bayes model on the 10 million word Bijankhan corpus \cite{bijankhan2011lessons} with the IOB tagging scheme to find word boundaries, achieving an F\textsubscript{1}-score of 96.5\%.
As for the latest work, \newcite{panahandeh2019correction} uses an N-gram language model and a rule-based method to correct space and ZWNJ between compound words, respectively. They report an F\textsubscript{1}-score of 81.94\% for space correction.

\section{Data}
\label{sec:data}

The 10 million word Bijankhan corpus \cite{bijankhan2011lessons} was used in this research, which is a cleanly tokenized corpus with fine-grained part-of-speech tags (which were not used here). This corpus contains 10,437,194 words or 38,971,131 characters. We chose this corpus since (a) their word segmentation is clean and their approach is practical, (b) the size of the corpus is enormous ($\sim$39M characters), and (c) the corpus comprises many different topics, including news articles, literary prose and poems, informal dialogues, etc. All white spaces in a single token were converted to ZWNJs and the characters in the corpus were normalized as well. Then, the corpus was split into three parts: the first 10\% for testing, the second 10\% for validation, and the rest for training. We also collected a test corpus of 500 sentences from twitter, news broadcasting websites, and discussion forums to have real-world data with real white space and ZWNJ errors. This data was deliberately meant to be a difficult test corpus with numerous extreme cases (and also easy and completely correct ones). This corpus contains 16,574 words or 93,355 characters.

\section{Methodology}
\label{sec:meth}

We see the word segmentation correction and the ZWNJ recognition tasks as one problem and train a single model to perform these two tasks jointly. We approached the task as a sequence labeling problem, i.e., mapping each character to a tag space of size $3$. The tag is \texttt{0} when there is no white space or ZWNJ, \texttt{1} when there is white space, and \texttt{2} when there is ZWNJ after the corresponding character. Two types of models were trained for this purpose:

\begin{enumerate}
    \item {\bf CRF}: a conditional random field (CRF) model \cite{lafferty2001conditional} implemented using \texttt{sklearn-crfsuite} \cite{sklearncrfsuite,CRFsuite}, with the input features of the focus, 5 previous and 5 following characters, and four character-based Boolean features indicating whether the focus character \texttt{is\_first} and \texttt{is\_last} character of the sentence, and also if the character \texttt{is\_joiner}, and whether it \texttt{is\_digit}. All the white space and ZWNJ characters were stripped from the input texts. The L1 and L2 regularization coefficients were set to $0.1$ and the max iteration argument to 100.
    
    \item {\bf BERT}: The main bidirectional encoder representations from Transformers (BERT) model \cite{devlin2018bert}, plus a fully-connected network mapping to the tag space. The learning rate was set to $2e-5$ and the batch size to 10. Adam \cite{kingma2014adam} was used for optimizing the weights with cross-entropy as the loss function. As for the pre-trained weights, the \texttt{multilingual cased} model was used. We have followed the recommended settings for sequence labeling, which is to calculate loss only on the first part of each tokenized word. The implementation was done using PyTorch \cite{NEURIPS2019_9015} and HuggingFace's Transformers \cite{Wolf2019HuggingFacesTS} libraries. The input texts were fed into the model in two different settings:
    
    \begin{enumerate}
        \item All the white space and ZWNJ characters were stripped from the input texts, similar to the CRF model above. The model has to figure out the position of white space and ZWNJ characters in sentences from scratch. 
        We refer to this model as BERT\textsubscript{a}.
        
        \item The white space and ZWNJ characters were kept, but some noise was introduced to the input data in the following manner for each sentence in Bijankhan corpus:
        
        \begin{enumerate}
            \item $r_1 \times l$ of the ZWNJs were changed to white space characters, where $l$ is the length of the sentence (in characters) and $r_1$ is a uniform random variable on $(0., .15)$.
            
            \item $r_2 \times l$ of the white spaces after non-joiner characters were removed, where $l$ is the length of the sentence (in characters) and $r_2$ is a uniform random variable on $(0., .2)$.
            
            \item $r_3 \times l$ of the remaining characters were changed in the following manner, where $l$ is the length of the sentence (in characters) and $r_3$ is a uniform random variable on $(0., .05)$:
            
            \begin{enumerate}
                \item $replace(c, rand(null, {\scriptstyle \mathit{ZWNJ}}))$, where the randomly chosen character $c$ was white space;
                \item $replace(c, rand(null, space))$, where the randomly chosen character $c$ was ZWNJ;
                \item $replace(c, concat(c, rand({\scriptstyle \mathit{ZWNJ}}, space)))$, where the randomly chosen character $c$ was followed neither by a white space nor a ZWNJ, otherwise, remove the following white space or ZWNJ.
            \end{enumerate}
            
        \end{enumerate}
        
        The noise ranges were chosen based on our observation of real-world errors; hence this scenario is closer to real-world situations. The corresponding output tags of the model for the input white space and ZWNJ character were masked and ignored in calculating the loss and performance measures. We refer to this model as BERT\textsubscript{b}.
    \end{enumerate}
\end{enumerate}

We also experimented with Parsivar tool, introduced in \cref{sec:rel}, to rectify white spaces and ZWNJs using its \texttt{Normalizer} (with the \texttt{statistical\_space\_correction} argument set to \texttt{True}) and \texttt{SpellCheck} modules. As the latter module might add or remove characters other than white space and ZWNJ, the differences between the two strings were found using Python's \texttt{difflib} library, and placeholder characters were added where needed to make the strings the same length and make their comparison possible. This work's performance measures are precision, recall, and F\textsubscript{1}-score for each class and macro-averaged F\textsubscript{1}-score of all of them.

\section{Results and Analysis}
\label{sec:res}

The results of the abovementioned methods on the test set of Bijankhan corpus and the 500 sentence corpus are shown in Table \ref{tab:results}. The BERT models outperform the other methods by a large margin, BERT\textsubscript{b} standing on the top by 
1.33\% F\textsubscript{1}-score more than BERT\textsubscript{a}, 
28.65\% more than the CRF, 
26.73\% more than Parsivar, and
9.55\% more than the baseline
on the 500 sentence corpus.
The baseline simply indicates the correctness of word segmentation and ZWNJs in the 500 test corpus (i.e., naturally occurring errors in the real-world data) when compared to its corrected pair. 
The results also show that Parsivar and the CRF model not only do not increase word segmentation and ZWNJs correctness, but also add more errors to the data. 

\begin{table}[h]
\centering
\begin{tabular}{ccccc|cccc}
\Xcline{2-9}{2\arrayrulewidth}
& \multicolumn{4}{c|}{\bf Bijankhan Test Set} & \multicolumn{4}{c}{\bf 500 Sentence Corpus} \\
\cline{2-9}
& \multicolumn{3}{c}{\bf Class F\textsubscript{1}} & \multirow{2}{*}{\bf Avg. F\textsubscript{1}} & \multicolumn{3}{c}{\bf Class F\textsubscript{1}} & \multirow{2}{*}{\bf Avg. F\textsubscript{1}} \\
& \bf \texttt{0} & \bf \texttt{1} & \bf \texttt{2} & & \bf \texttt{0} & \bf \texttt{1} & \bf \texttt{2} &  \\
\cline{1-9}
\bf Baseline & & & & & 0.9823 & 0.9336 & 0.5697 & 0.8285 \\
\bf Parsivar & & & & & 0.9561 & 0.7785 & 0.2355 & 0.6567 \\
\bf CRF & 0.8923 & 0.6369 & 0.5664 & 0.6985 & 0.8826 & 0.6080 & 0.4217 & 0.6375 \\
\bf BERT\textsubscript{a} & 0.9937 & 0.9779 & 0.9286 & 0.9667 & 0.9832 & 0.9488 & 0.8000 & 0.9107\\
\bf BERT\textsubscript{b} & \bf 0.9963 & \bf 0.9886 & \bf 0.9593 & \bf 0.9814 & \bf 0.9856 & \bf 0.9592 & \bf 0.8272 & \bf 0.9240\\
\Xcline{1-9}{2\arrayrulewidth}
\end{tabular}
\caption{\label{tab:results} The F\textsubscript{1}-score for each class and the macro-averaged F\textsubscript{1}-scores of all the classes for each of the methods on Bijankhan test set and the 500 sentence corpus. The best result in each column is in bold.}
\end{table}

The errors in BERT\textsubscript{a} and BERT\textsubscript{b} are more or less similar and can be categorized into the following groups: (a) ZWNJ before an intrusive \emph{y} between a vowel and an \emph{ezafe} (e.g., \emph{x\={a}ney[e]} ``[the] house of"), which is simply because this writing style is not used in Bijankhan corpus (and is also fixable as words with \emph{ezafe} are labeled as \texttt{GEN} in Bijankhan corpus); (b) underscores and hash characters in hashtags, which is again because the model has not seen it in the training data; (c) out of vocabulary words, such as \emph{tu'iter} ``twitter" and \emph{d\={a}m\={a}\v{s}} ``damash (a soccer team name)"; (d) digits occasionally sticking to the word before or after them; (e) informal words, such as the clitic \emph{e} ``is", which is the short form of \emph{ast} ``is"; (f) words or phrases with different syntactic roles, such as \emph{xarid\={a}ri$|$\v{s}ode} ``bought (adjective)" and \emph{xarid\={a}ri \v{s}ode} ``bought (verb)"; (g) typos and uncommon spellings. The confusion matrix also reveals that ZWNJ is mostly mistaken with the \texttt{0} class, and not with the \texttt{1} class, which results in joiner characters connecting into a ligature and make the text difficult to read.
There are also some cases which can not strictly be counted as errors, but different writing styles. Unfortunately, our performance measure does not account for these, more or less, common cases. Some examples are \emph{pul$|$d\={a}r} ``rich", \emph{doruq$|$gu} ``liar", and \emph{\v{s}aff\={a}f$|$s\={a}zi} ``clarification". All in all, adding more training samples with the abovementioned features would most probably solve many of the mentioned error groups.

\section{Conclusion and Future Work}
In this paper, we experimented with different methods to tackle word segmentation correction and zero-width non-joiner recognition problems in Persian. The results on our collected data show BERT outperforming other methods, and the error analysis indicates that it would be relatively easy to increase the performance and pave the way for a practical and effective preprocessing tool. Future work could focus on collecting more informal Persian data, more diverse topics, and more modern registers of the language, to further improve this work's results. There are also some techniques in the work mentioned in \cref{sec:rel}, e.g., multi-task learning (say, for the model to learn the difference between joiner and non-joiner characters as an auxiliary task), that could be used in Persian as well. Covering different writing styles in the training data would also be helpful, e.g., changing some of the ending \emph{e[ye]} clitics in Bijankhan corpus, when the word is tagged as \texttt{GEN}, to \emph{ey[e]}, as discussed in \cref{sec:res}.

\bibliographystyle{coling}
\bibliography{coling2020.bib}

\end{document}